\documentclass[conference]{IEEEtran}
\IEEEoverridecommandlockouts
\usepackage[T1]{fontenc}
\usepackage{cite}
\usepackage{amsmath}
\usepackage{amsthm}
\usepackage{amssymb}
\usepackage{amsfonts}
\usepackage{graphicx}
\usepackage{epstopdf}
\usepackage{textcomp}
\usepackage{xcolor}
\usepackage{multirow}
\usepackage[normalem]{ulem}
\useunder{\uline}{\ul}{}
\usepackage{subfigure}

\newcommand{\nosection}[1]{\vspace{2pt}\noindent\textbf{#1.}}
\newcommand{\modelname}{\texttt{DGDFEM}}
\newcommand{\gnnname}{\texttt{HLGCN}}

\newtheorem{Proposition}{Proposition}

\def\BibTeX{{\rm B\kern-.05em{\sc i\kern-.025em b}\kern-.08em
    T\kern-.1667em\lower.7ex\hbox{E}\kern-.125emX}}
\begin{document}

\title{Freshness or Accuracy, Why Not Both? Addressing Delayed Feedback via Dynamic Graph Neural Networks\thanks{$^*$Corresponding author.}}

\author{  Xiaolin Zheng{$^{1}$}, Zhongyu Wang{$^{1}$}, Chaochao Chen{$^{1,*}$}, Feng Zhu{$^{2}$}, Jiashu Qian{$^{1}$} \\
{$^{1}$}College of Computer Science and Technology, Zhejiang University, Hangzhou, China \\
{$^{2}$}Ant Group, Hangzhou, China \\
\{xlzheng, iwzy7071, zjuccc\}@zju.edu.cn, zhufeng.zhu@antgroup.com, iqjs0124@gmail.com
}

\maketitle

\begin{abstract}
The \textit{delayed feedback} problem is one of the most pressing challenges in predicting the conversion rate since users' conversions are always delayed in online commercial systems.
Although new data are beneficial for continuous training, without complete feedback information, i.e., conversion labels, training algorithms may suffer from overwhelming \textit{fake negatives}. 
Existing methods tend to use multitask learning or design data pipelines to solve the delayed feedback problem.
However, these methods have a trade-off between data freshness and label accuracy.
In this paper, we propose \textbf{D}elayed \textbf{Fe}edback \textbf{M}odeling by \textbf{D}ynamic \textbf{G}raph Neural Network (\modelname).
It includes three stages, i.e., preparing a data pipeline, building a dynamic graph, and training a CVR prediction model.
In the model training, we propose a novel graph convolutional method named \gnnname, which leverages both high-pass and low-pass filters to deal with conversion and non-conversion relationships.
The proposed method achieves both data freshness and label accuracy.
We conduct extensive experiments on three industry datasets, which validate the consistent superiority of our method.
\end{abstract}


\section{Introduction}
\label{sec:intro}
Conversion rate (CVR)-related problems are fundamental in the setting of E-commerce. 
Most CVR-related methods focus on the CVR prediction~\cite{CVR_SUPPORT_0, CVR_SUPPORT_1, CVR_SUPPORT_2}, which aims to estimate the probabilities of specific user behaviors, e.g., buying recommended items.
However, if conversions are not collected in time, i.e., the user feedback is delayed, it is difficult to predict CVR accurately.

\begin{figure}[t]
\centering

\subfigure[Delayed feedback problem.]{
\includegraphics[width=4cm]{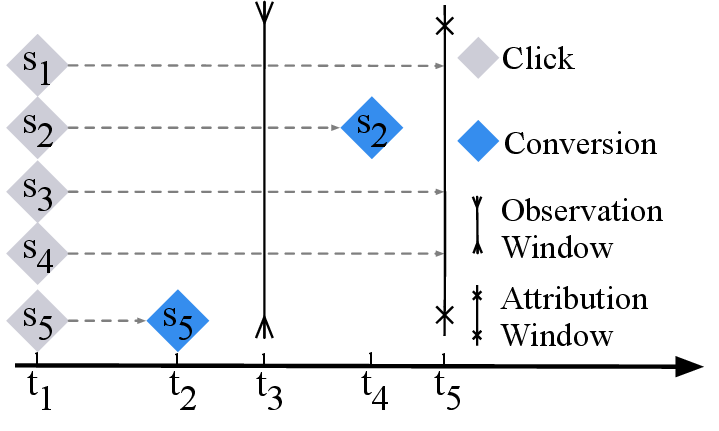}
\label{Figure_Intro_Delayed_Feedback_Problem}
}
\quad
\subfigure[Percentages of conversions.]{
\includegraphics[width=3.5cm]{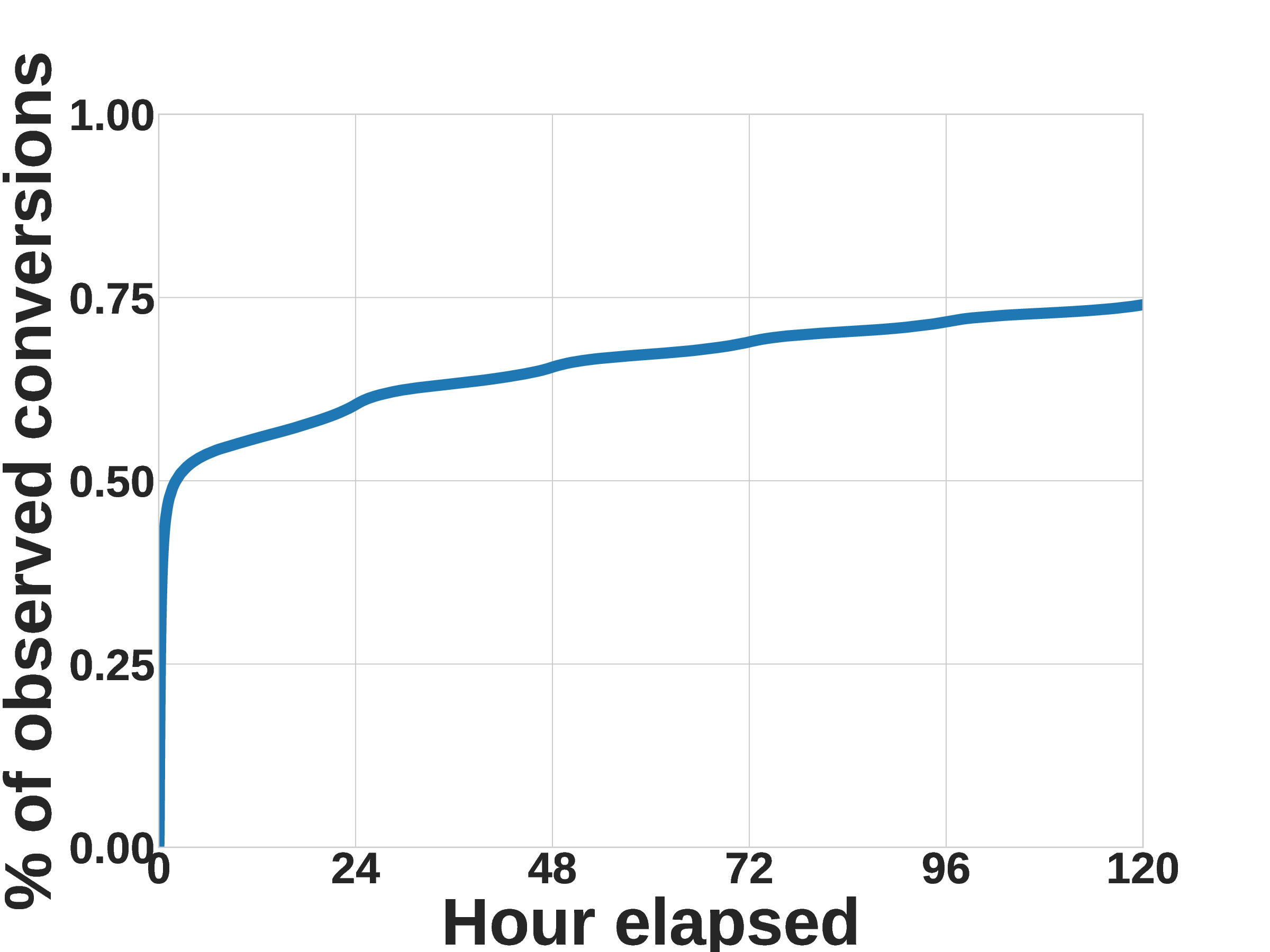}
\label{Figure_Intro_Delayed_Feedback_Distribution}
}
\caption{Introduction of the delayed feedback problem.
}
\end{figure}

Here, we take two examples to introduce the delayed feedback problem. 
Fig.~\ref{Figure_Intro_Delayed_Feedback_Problem} illustrates why this problem exists.
Without loss of generality, suppose these clicks occur at time $t_{1}$.
For most clicks, there are no conversions, e.g., sample $s_{1}$, $s_{3}$, and $s_{4}$.
For the rest, their corresponding conversions are delayed for different times, e.g., sample $s_{2}$ and $s_{5}$.
When setting an observation window at time $t_{3}$, we can only observe the conversion of sample $s_{5}$ at time $t_{2}$.
To distinguish whether these samples will convert, we should postpone the observation window until the \textit{attribution window} ends at time $t_{5}$. 
An attribution window is a period of time in which we can claim that a conversion is led by a click.
As a result, the delivery of training samples is significantly delayed. 
Although the delay feedback problem exists, most online commercial systems train their CVR prediction models in real-time data streams to capture data distribution changes. 
Fig.~\ref{Figure_Intro_Delayed_Feedback_Distribution} illustrates the importance of delayed feedback modeling with a real-world dataset, i.e., Criteo Sponsored Search Traffic~\cite{CRITEO2018_DATASET}.
From it, we can observe half of the conversions in several hours but need to wait a couple of days or even longer to observe the rest. 
The label accuracy of training samples increases with more observed conversions while data freshness decreases.
However, both data freshness and label accuracy are necessary for online commercial systems.
Training with new data enhances real-time response capability for online systems.
Training with accurate sample labels enhances model performance.

Towards addressing the delayed feedback problem, we can divide existing methods into two types, i.e., \textit{mutitask learning} and \textit{designing data pipelines}. 
The former~\cite{DFM, PARAM_DFM, JD_ATTENTION, DLADF, MTDFM} uses additional models to assist in learning the CVR estimation.
But it waits a long time to collect all conversions, leading to recommendation performance degradation.
The latter~\cite{TWITTER_FNW, FSIW, ESDFM, DEFER, DEFUSE, TENCENT_PROPHET} designs new data pipelines for CVR prediction models.
But it faces a trade-off between data freshness and label accuracy simultaneously. 
In summary, the above models cannot solve the delayed feedback problem.

To address the delayed feedback problem successfully, we propose \textbf{D}elayed \textbf{Fe}edback \textbf{M}odeling by \textbf{D}ynamic \textbf{G}raph Neural Network (\modelname).
%
The framework of \modelname~ includes three stages, i.e., preparing a data pipeline (Stage 1), building a dynamic graph (Stage 2), and training a CVR prediction model (Stage 3). 
\textbf{In Stage 1}, we design a novel data pipeline that achieves the best data freshness. 
To capture data freshness, we deliver each sample, i.e., a user-item click, as soon as it appears, even when it is \textit{unlabeled}. 
These delivered unlabeled samples are used to build the dynamic graph. 
To achieve high label accuracy, we set a time window to wait for each sample's conversion. 
When the time window ends, we mark the sample as either positive or negative and deliver it again for the following two stages. 
Finally, after fake negatives are converted, we calibrate them as positive and deliver them one more time. 
\textbf{In Stage 2}, we build a dynamic graph to facilitate CVR prediction by capturing frequent changes in data distribution.
We build the dynamic graph upon a sequence of chronologically delivered samples. 
In the graph, nodes are the users and items from samples, and edge attributes come from sample labels.
\textbf{In Stage 3}, we take users and items in the delivered samples as seeds to sample multi-hop neighbors from the dynamic graph.
To deal with conversion and non-conversion relationships, we propose a novel graph convolutional method, namely \gnnname, which aggregates features through \textit{high-pass} and \textit{low-pass} filters.
High-pass filters capture node differences among non-conversion relationships.
Low-pass filters retain node commonalities among conversion relationships. 
To alleviate the noises introduced by fake negatives, we further propose \textit{distribution debias} and prove its effectiveness theoretically.
Our method achieves both data freshness and label accuracy.

We summarize the main contributions of this paper as follows:
\textbf{(1)} We propose a novel delayed feedback modeling framework that highlights label accuracy with the best data freshness.
\textbf{(2)} We propose the first dynamic graph neural network to solve the delayed feedback problem, which can capture data distribution changes and hence facilitates CVR predictions.
\textbf{(3)} We conduct extensive experiments on three large-scale industry datasets. 
Consistent superiority validates the success of our proposed framework.

\section{Related Work}
\subsection{Delayed Feedback Modelling}
In solving the delayed feedback problem, we can divide existing methods into two types, i.e., multitask learning and designing data pipelines.

\nosection{Multitask Learning} 
This method uses additional models to assist in learning the CVR estimation.
DFM~\cite{DFM} designs two models for estimating CVR and delay time.
One successor of DFM, i.e., NoDeF~\cite{PARAM_DFM}, proposes fitting the delay time distribution by neural networks.
Another successor~\cite{JD_ATTENTION} extracts pre-trained embeddings from impressions and clicks to predict CVR and leverages post-click information to estimate delay time.
DLA-DF~\cite{DLADF} trains a CVR predictor and propensity score estimator based on a dual-learning algorithm.
MTDFM~\cite{MTDFM} trains the actual CVR network by simultaneously optimizing the observed conversion rate and non-delayed positive rate.
These methods wait a long time to collect all conversions, leading to recommendation performance degradation.
In contrast, our method delivers samples as soon as they appear, achieving the best data freshness.

\nosection{Designing Data Pipelines} 
This method designs new data pipelines to change the way data is delivered.
FNW and FNC~\cite{TWITTER_FNW} deliver all samples instantly as negative and duplicate them as positive when their conversions occur.
But these two methods introduce numerous fake negatives into model training.
FSIW~\cite{FSIW} waits a long time for conversions.
But it uses stale training data and does not allow the correction, even when the conversion occurs afterward.
ES-DFM~\cite{ESDFM} sets a short time window to wait for conversions and rectifies these fake negatives.
Based on ES-DFM, DEFER~\cite{DEFER} and DEFUSE~\cite{DEFUSE} further duplicate and ingest real negatives for training.
But ES-DFM, DEFER, and DEFUSE cannot simultaneously achieve label accuracy and data freshness.
As the time window extends, these three methods observe more conversion feedback, i.e., label accuracy becomes higher, but data freshness gets lower.
FTP~\cite{TENCENT_PROPHET} constructs an aggregation policy on top of multi-task predictions, where each task captures the feedback pattern during different periods.
But some tasks in FTP are trained with accurate but stale data, while others are trained with new but inaccurate data.
These methods face a trade-off between data freshness and label accuracy.
In contrast, our method achieves both data freshness and label accuracy.
%

\subsection{Dynamic Graph Neural Networks}
In dynamic graph neural networks, we can divide existing methods into two types, i.e., discrete-time dynamic graphs and continuous-time dynamic graphs.
\textit{Discrete-time dynamic graph} represents a dynamic graph as a series of static graph snapshots at different time intervals.
DySAT~\cite{DYSAT} computes node representations through joint self-attention within and across the snapshots.
DGNF~\cite{DGNF} captures the missing dynamic propagation information in graph snapshots by dynamic propagation.
TCDGE~\cite{tcdge} co-trains a linear regressor to embed edge timespans and infers a common latent space for all snapshot graphs.
\textit{Continuous-time dynamic graph} represents a dynamic graph as event sequences, e.g., creating nodes and adding edges. 
DyGNN~\cite{Stream_GNN} updates node states involved in an interaction and propagates the updates to neighbors. 
GNN-DSR~\cite{GNNDSR} models the short-term dynamic and long-term static representations of users and items and aggregate neighbor influences.
HDGCN~\cite{hdgcn} explores dynamic features based on a content-propagation heterogeneous graph.
TGN~\cite{TGN} combines different graph-based operators with memory modules to aggregate temporal-topological neighborhood features and learns time-featured interactions.
All these methods leverage stale or even wrong interactions in delayed feedback scenarios since the samples used to build the graph are delayed and mislabeled.
In contrast, our method provides the newest data with accurate labels for the dynamic graph and proposes \gnnname~to deal with conversion and non-conversion relationships in the graph. 
\section{Method}
In this section, we present the framework of \modelname~followed by its main stages in detail.
\modelname~has three stages, i.e.,  preparing a data pipeline, building a dynamic graph, and training a CVR prediction model. 
In the first stage, we prepare a data pipeline that achieves high label accuracy with the best data freshness.
In the second stage, we build a dynamic graph, which captures data distribution changes.
In the third stage, we propose a graph convolutional method named \gnnname~to deal with conversion and non-conversion relationships.
Besides, we propose \textit{distribution debias} to alleviate the noises from fake negatives.
\modelname~achieves both data freshness and label accuracy, which solves the delayed feedback problem.

\begin{figure}[t]
\centering
\includegraphics[width=\linewidth]{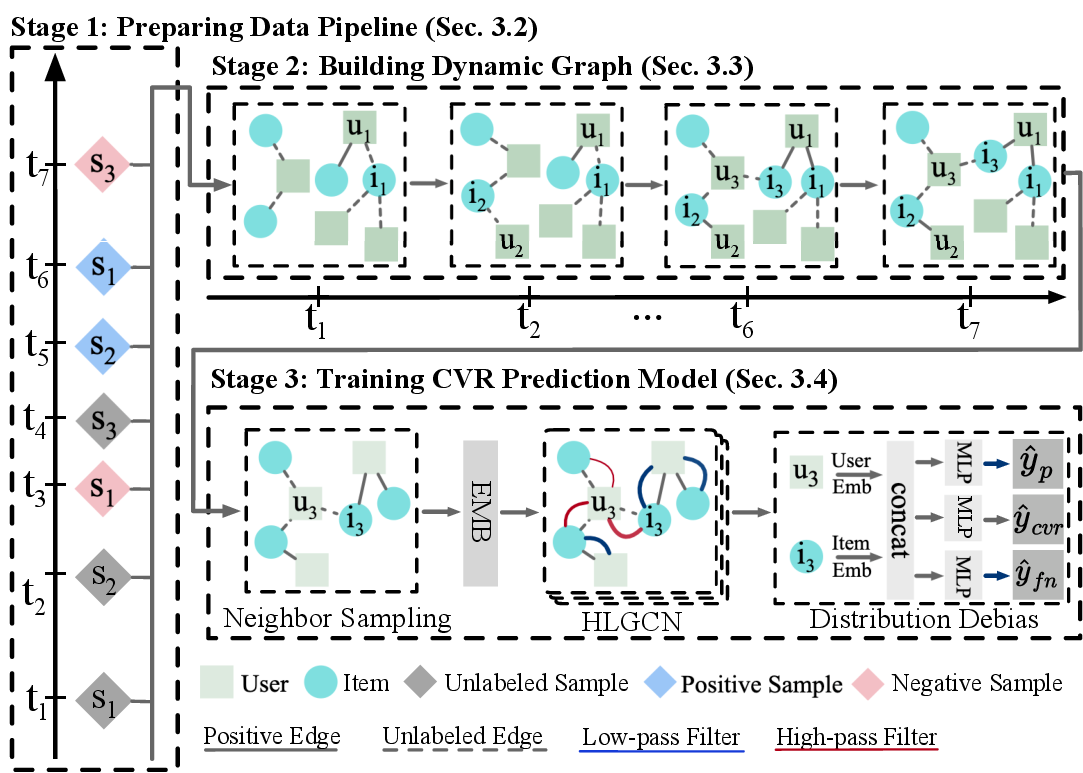}
\caption{
The proposed \modelname~framework.
The solid arrowed lines present the framework workflow.
The dashed arrowed lines separate the changes in the graph at different times. 
}
\label{Figure_Architecture_DGDFM_Overall_Architecture}
\end{figure}

\subsection{Preparing Data Pipeline}
\label{Preparing Data Pipeline}
\nosection{Preliminary}
There are three possible types of samples in the data pipeline of \modelname, namely \textit{real negative}, \textit{fake negative}, and \textit{positive}.
For ease of description, we let $t_{d}$ denote the delay time of conversion feedback, $l_{w}$ denote the time window, and $l_{a}$ denote the attribution window length.
We define three possible sample types as follows.
\textbf{(1)} \textit{Fake negative} ($l_{w}$ $\leq$ $t_{d}$ $\leq$ $l_{a}$) is the sample whose conversion occurs beyond the time window.
It is incorrectly delivered as negative at first.
\textbf{(2)} \textit{Positive} ($t_{d}$ < $l_{w}$) is the sample whose conversion occurs within the time window.
\textbf{(3)} \textit{Real negative} ($t_{d}$ > $l_{a}$) is the sample that does not convert eventually.
We set $t_{d}=\infty$ for these samples.

From the above definition, we observe that time window length $l_{w}$ determines the proportions of fake negatives and positives.
When we set $l_{w} \to 0$, we transform all positives into fake negatives.
When we set $l_{w} \to l_{a}$, we eliminate all fake negatives.

\begin{figure}[t]
\centering
\includegraphics[width=\linewidth]{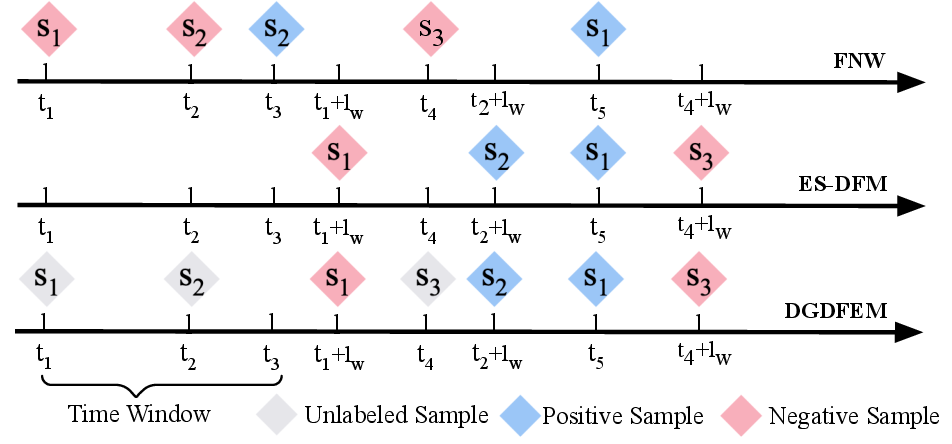}
\caption{Data pipelines of FNW, ES-DFM, and~\modelname.}
\label{Figure_DGDFM_Data_Pipeline_Difference}
\end{figure}

\nosection{Our Proposed Data Pipeline}
We describe our proposed data pipeline through an example illustrated in Fig.~\ref{Figure_DGDFM_Data_Pipeline_Difference}. 
Suppose there are three samples, denoted as $s_{1}$, $s_{2}$, and $s_{3}$, representing `fake negative,' `positive,' and `real negative,' respectively.
We first deliver each sample, i.e., a user-item click, instantly into the data pipeline as soon as it appears in the commercial system, even when it is unlabeled.
We set a time window to wait for each sample's conversion and deal with different sample types in various ways.
\textbf{(1)} For fake negative $s_{1}$, which arrives at time $t_{1}$, we set a time window between time $t_{1}$ and $t_{1}+l_{w}$.
When the window ends at time $t_{1}+l_{w}$, we mark it as negative and deliver it for a second time as we do not observe the conversion feedback within the window.
Finally, we calibrate it as positive and deliver it one more time when its conversion occurs at time $t_{5}$.
\textbf{(2)} For positive $s_{2}$, which arrives at time $t_{2}$, we set a time window between time $t_{2}$ and $t_{2}+l_{w}$.
When the window ends at time $t_{2}+l_{w}$, we mark it as positive and deliver it again as its conversion occurs at time $t_{3}$, i.e., within the time window.
\textbf{(3)} For real negative $s_{3}$, which arrives at time $t_{4}$, when the time window ends at time $t_{4}+l_{w}$, we mark it as negative and deliver it.

We utilize these three sample types differently.
We deliver unlabeled samples to build the dynamic graph and negative samples.
to train the model.
We deliver positive samples to build the dynamic graph and train the model.

\nosection{Comparison with Existing Data Pipelines}
We further analyze the differences in the data pipelines between our method and the state-of-the-art methods, i.e., FNW~\cite{TWITTER_FNW} and ES-DFM~\cite{ESDFM} to demonstrate our motivations. 
\textbf{(1)} FNW marks each sample as negative when it appears and delivers it instantly for training the model, e.g., sample $s_{1}$ at time $t_{1}$.
This introduces numerous fake negatives into model training, which lowers label accuracy.
In contrast, \modelname~delivers each sample for model training after the time window ends.
There is sufficient time for most samples to convert, i.e., \modelname~achieves high label accuracy.
\textbf{(2)} ES-DFM marks all samples as either positive or negative and delivers each sample after its time window, e.g., sample $s_{1}$ at time $t_{1}+l_{w}$.
ES-DFM postpones the delivery for each sample until the end of the time window, which lowers data freshness. 
In contrast, \modelname~first delivers samples instantly to build a dynamic graph, achieving the best data freshness.
To this end, \modelname~enjoys high label accuracy with the best data freshness.

\subsection{Building the Dynamic Graph}
\label{Building Dynamic Graph}
\nosection{Motivation of the Dynamic Graph}
Conversions correlate with data distribution changes.
For example, when an item is on sale, users are more inclined to buy it.
As a result, conversions associated with this item are more likely to occur.

Leveraging a dynamic graph can better capture the data distribution changes, facilitating CVR prediction.
We can represent users and items in delivered samples as a graph.
In the graph, nodes are users and items, and edge attributes come from sample labels.
Our dynamic graph explicitly models data distribution changes as variations of node attributes and edges.

\nosection{Dynamic Graph Definition}
We build the dynamic graph upon a sequence of chronologically delivered samples, i.e., $\mathcal{G}=\{s(t_{1}),s(t_{2}),...\}$ at times $0 \leq t_{1} \leq t_{2} \leq \dots$.
Besides, we set $s=(u,i,y)$ where $u$ and $i$ are the user and item in sample $s$, and $y \in \{-1,0,1\}$ denotes the sample label, with $-1$ denoting the unlabeled one, $0$ denoting the negative one, and $1$ denoting the positive one.

\nosection{Building the Dynamic Graph}
Delivered samples are involved in \textit{node-wise events} and \textit{edge-wise events} to build the dynamic graph chronologically~\cite{TGN}. 
The former creates nodes or updates node features, and the latter adds or deletes edges.

We represent a node-wise event by $v_{j}(t)$, where $j$ is the node index and $v$ is the vector attribute associated with the sample.
This event \textit{only} occurs for unlabeled samples as they carry the latest data information.
In contrast, positive and negative samples are duplicates of previous unlabeled samples.
If node $j$ is not in $\mathcal{G}$, the event creates it along with its attributes. 
Otherwise, the event updates its feature.
Take Fig.~\ref{Figure_Architecture_DGDFM_Overall_Architecture} as an example.
unlabeled sample $s_{2}$ is involved in two node-wise events at time $t_{2}$, i.e., adding a new node $u_{2}$ to the graph and updating the feature of node $i_{2}$.

We represent an edge-wise event as $e_{uiy}(t)$, where $y$ is the sample label.
It occurs for unlabeled and positive samples, i.e., $s$ with $y \in \{-1, 1\}$ as negative samples ($s$ with $y = 0$) may be fake negative and represent wrong interactions.
We keep at most $m$ \textit{recent} edges for each node and delete the rest to improve efficiency.
In Fig.~\ref{Figure_Architecture_DGDFM_Overall_Architecture}, unlabeled sample $s_{1}$ is first involved in an edge-wise event to add an unlabeled edge between user $u_{1}$ and item $i_{1}$ at time $t_{1}$.
It is then involved in another edge-wise event to connect user $u_{1}$ and item $i_{1}$ with a positive edge when it becomes positive at time $t_{6}$.

After describing node-wise and edge-wise events, we define the dynamic graph at time $t$ as follows.
In time interval $T$, we denote temporal sets of nodes and edges as $\mathcal{V}_{T}=\{j:\exists v_{j}(t) \in \mathcal{G}, t \in T\}$ and $\mathcal{E}_{T}=\{(u,i,y):\exists e_{uiy}(t) \in \mathcal{G}, t \in T\}$.
We denote a snapshot of dynamic graph $\mathcal{G}$ at time $t$ as $\mathcal{G}_{t}=(\mathcal{V}_{[0, t]}, \mathcal{E}_{[0, t]})$.
We denote the neighbors of node $j$ at time $t$ as $\mathcal{N}_{j}(t) = \{m:(j,m) \in \mathcal{E}_{[0, t]}\}$ and the $k$-hop neighborhood as $\mathcal{N}_{j}^{k}(t)$.

\subsection{Training CVR Prediction Model}
\label{Training CVR Prediction Model}
This section introduces training a CVR prediction model in \modelname.
We use positive and negative samples for model training.
For each positive sample, the model training process occurs before updating the dynamic graph to prevent the model from leveraging future information.
We can further divide the whole training process into four steps.

\nosection{Neighbor Sampling} 
Online commercial systems require real-time services for recommendation and advertising~\cite{TWITTER_FNW, real_time}.
To balance model performance and response time, we take the user and item as seeds to sample $k$-hop neighbors from dynamic graph $\mathcal{G}_{t}$.
We use Fig.~\ref{Figure_Architecture_DGDFM_Overall_Architecture} to exemplify neighbor sampling with $k=2$.
For sample $s_{3}$ delivered at time $t_{7}$, its corresponding user $u_{3}$ and item $i_{3}$ are used to sample $2$-hop neighbors from the dynamic graph.
To simplify the description, we continue to use $\mathcal{G}_{t}$ to represent the sampled graph.

\nosection{Node Embedding} To describe nodes in the sampled graph, we embed them based on their attributes.
Node features consist of sparse and dense features.
The former is embedded through embedding matrices, whereas the latter is scaled into $[0, 1]$ through min-max normalization.
We use a multilayer perceptron (MLP) model to map them into the same dimension, as user and item features do not necessarily share the same dimension size.
We define the embedding process as,
\begin{equation}
e^{(0)}_{j} = \mathrm{MLP}([x_{j,1},x_{j,2},...,x_{j,m},...]),\nonumber
\end{equation}
where $x_{j,m}$ denotes the $m$-th processed feature of node $j \in \mathcal{N}_{u}^{K}(t) \bigcup \mathcal{N}_{i}^{K}(t)$,
$[\cdot]$ denotes the concatenation function,
and $e_{j}^{(0)}$ denotes the initial embedding of node $j$.

\nosection{\gnnname} After processing node embeddings, we propose a novel graph convolutional method named \gnnname~to gather features from neighbors.
In our setting, a successful graph convolutional method should deal with non-conversion and conversion relationships in the dynamic graph.
Traditional graph convolutional methods~\cite{GCN, GAT} can only handle conversion relationships as they use low-pass filters to retain node commonalities.

Our proposed \gnnname~uses both \textit{high-pass} and \textit{low-pass} filters. 
The former captures node differences among non-conversion relationships, whereas the latter retains node commonalities among conversion relationships.
It aggregates features from neighbors, as shown in~\eqref{Equation_Message_Passing}.
We introduce its implementation details later in Section~\ref{subsec:AHLPF}. 
\begin{equation}
e_{u}^{(l)} = g(e_{u}^{(0)}, e_{u}^{(l-1)}, \{e_{i}^{(l-1)} \mid i \in \mathcal{N}_{u}(t) \}),
\label{Equation_Message_Passing}
\end{equation}
where $e_{u}^{(l)}$ is the aggregated embedding of node $u$ from layer $l \in \{1 \dots k\}$, and $g(\cdot)$ denotes the graph convolution.

\nosection{Distribution Debias}
Although we calibrate fake negatives in our data pipeline, these samples shift actual data distribution as fake negatives are duplicated.
The gap between actual and biased data distribution results in training noises.
Here, we leverage the importance sampling~\cite{importance_sampling} to alleviate the training noises.
Compared with ESDFM~\cite{ESDFM}, we predict sample probabilities of being positive and fake negative to assist in learning CVR.
We define the prediction process as,
\begin{equation}
\begin{aligned}
\hat{y}_{p} &=\mathrm{Sigmoid}(\mathrm{MLP}_{p}([e_{u}^{(k)},e_{i}^{(k)}])),\\ 
\hat{y}_{fn} &=\mathrm{Sigmoid}(\mathrm{MLP}_{fn}([e_{u}^{(k)},e_{i}^{(k)}])),\\
\hat{y}_{cvr} &=\mathrm{Sigmoid}(\mathrm{MLP}_{cvr}([e_{u}^{(k)},e_{i}^{(k)}])).\nonumber
\end{aligned}
\label{equ_calculating_three_equations}
\end{equation}

When updating the CVR prediction model, we take $\hat{y}_{fn}$ and $\hat{y}_{p}$ as values, cut down their backpropagation, and update these two prediction models separately.
We optimize the CVR prediction model by minimizing the loss as, 
\begin{equation}
\begin{aligned}
\mathcal{L} = -\sum_{(u, i, y_{ui}) \in \mathcal{S}}(y_{ui}(1+\hat{y}_{fn})log(\hat{y}_{cvr})\\+(1-y_{ui})(1+\hat{y}_{fn})(1-\hat{y}_{p}-\hat{y}_{fn})log(1-\hat{y}_{cvr})),
\end{aligned}
\label{equ_loss_function}
\end{equation}
where $\mathcal{S}$ is a set of positive and negative samples drawn from the data pipeline,
and $y_{ui}$ is the sample label, with $y_{ui} = 1$ denoting a positive sample and $y_{ui} = 0$ denoting a negative one.

\begin{Proposition}
The distribution debias can correct distribution shifts.
\label{proposition_distribution_alignment}
\end{Proposition}
Proposition~\ref{proposition_distribution_alignment} argues that our distribution debias alleviates training noises resulting from the fake negatives, which boosts CVR prediction accuracy.
We give its detailed proof in Appendix~\ref{appendix_proof_proposition1}.

\subsection{\gnnname}
\label{subsec:AHLPF}
\begin{figure}[t]
\centering
\includegraphics[width=\linewidth]{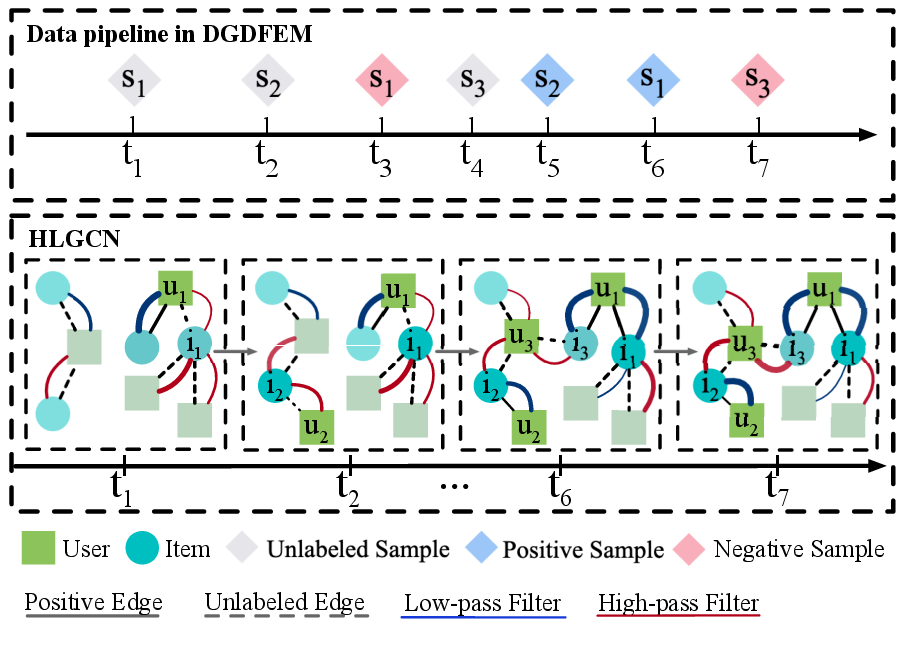}
\caption{
An illustration of the used filter for each edge in \gnnname.
The shades of node colors in the dynamic graph reflect feature changes over time.
The filter color reflects the change in filter type, and the filter thickness demonstrates the weight over time.
}
\label{fig:AHLPF}
\end{figure}

We have provided our motivation for \gnnname~in Section~\ref{Training CVR Prediction Model} that it should be able to deal with non-conversion and conversion relationships in the dynamic graph.
Here, we introduce its technical details.
\gnnname~calculates user preferences to determine each edge's filter type and weight as conversions only depend on user preferences~\cite{JD_ATTENTION}.
It then uses appropriate filters to exploit both types of relationships and aggregates neighbor features.
Fig.~\ref{fig:AHLPF} illustrates the aggregation process of \gnnname.
It first uses a high-pass filter for the edge between user $u_{1}$ and item $i_{1}$ at time $t_{1}$ and then replaces it with a low-pass filter at time $t_{6}$.
This is because our data pipeline delivers sample $s_{1}$ as positive at time $t_{6}$.
The edge between user $u_{1}$ and item $i_{1}$ changes from unlabeled to positive.
In other words, the preference of user $u_{1}$ on item $i_{1}$ is maximized at time $t_{6}$.

\nosection{Filter Definition}
A high-pass filter $F_{H}$ amplifies high-frequency signals, while a low-pass filter $F_{L}$ magnifies low-frequency signals~\cite{high_low_filter_gnn}.
We state the definitions filter $F_{H}$ and $F_{L}$ as,
\begin{equation}
\begin{aligned}
F_{H} = \epsilon I - D_{t}^{-1/2}A_{t}D_{t}^{-1/2},\\
F_{L} = \epsilon I + D_{t}^{-1/2}A_{t}D_{t}^{-1/2},\nonumber
\end{aligned}
\end{equation}
where $A_{t}$ denotes adjacency matrix of $\mathcal{G}_{t}$, $D_{t}$ denotes diagonal degree matrix,
$I$ denotes the identity matrix,
and $\epsilon$ is a hyper-parameter determining the retention proportion of a node's initial embedding.

We use $F_{H}$ to capture node difference and $F_{L}$ to retain node commonality.
We define their aggregation process as,
\begin{equation}
\begin{aligned}
\hat{e}_{u}^{(l)} &= F_{H} \cdot H_{t}= \epsilon \cdot e_{u}^{(0)} - \sum_{i \in \mathcal{N}_{u}(t)} e_{i}^{(l-1)}/\sqrt{d_{u}(t)d_{i}(t)},\\
\check{e}_{u}^{(l)} &= F_{L} \cdot H_{t}= \epsilon \cdot e_{u}^{(0)} + \sum_{i \in \mathcal{N}_{u}(t)} e_{i}^{(l-1)}/\sqrt{d_{u}(t)d_{i}(t)},
\end{aligned}
\label{equ: agg_high_pass_low_pass_filter}
\end{equation}
where $H_{t}$ denotes node embeddings and $d_{u}(t)$ denotes the degree of node $u$.
Besides, $\hat{e}_{u}^{(l)}$ and $\check{e}_{u}^{(l)}$ are aggregated embeddings of node $u$ at the $l$-th layer through $F_{H}$ and $F_{L}$, respectively.

\nosection{Aggregating Features from Neighbors}
Since we sample $k$-hop neighbors from $\mathcal{G}_{t}$, we use $k$ layers of \gnnname~to aggregate features from neighbors.
By combining~\eqref{Equation_Message_Passing} and~\eqref{equ: agg_high_pass_low_pass_filter}, we define its graph convolution $g(\cdot)$ as,
\begin{equation}
\begin{aligned}
e_{u}^{(l)} &= g(e_{u}^{(0)}, e_{u}^{(l-1)}, \{e_{i}^{(l-1)} \mid i \in \mathcal{N}_{u}(t)\})\\
&= \hat{w}\cdot \hat{e}_{u}^{(l)} + \check{w}\cdot \check{e}_{u}^{(l)}\\
&= \epsilon(\hat{w}_{ui}+\check{w}_{ui}) \cdot e_{u}^{(0)} \\
&+ \sum_{i \in \mathcal{N}_{u}(t)}(\hat{w}_{ui}-\check{w}_{ui})\cdot e_{i}^{(l-1)}/\sqrt{d_{u}(t)d_{i}(t)}\\
&= \epsilon \cdot e_{u}^{(0)} + \sum_{i \in \mathcal{N}_{u}(t)}p_{ui}\cdot e_{i}^{(l-1)}/\sqrt{d_{u}(t)d_{i}(t)},
\end{aligned}
\label{Equation_AHLPF}
\end{equation}
where $\hat{w}_{ui}$ and $\check{w}_{ui}$ denote the weight of the high-pass and low-pass filter for edge $e_{uiy}(t)$.
We set $\hat{w}_{ui}+\check{w}_{ui}=1$ as these two types of filters share the same processing on initial node embeddings.
We let user preference for the item $p_{ui}$ determine the types and weight of the used filters, i.e., $\check{w}_{ui}-\hat{w}_{ui}=p_{ui}$.

The above definition enables \gnnname~to use appropriate filters to exploit conversion and non-conversion relationships.
\textbf{(1)} When $p_{ui}<0$, it uses $F_{H}$ to capture node difference.
The weight of $F_{H}$ increases as $p_{ui}$ becomes lower.
\textbf{(2)} When $p_{ui}>0$, it uses $F_{L}$ to retain node commonality.
The weight of $F_{L}$ increases as $p_{ui}$ increases.
\textbf{(3)} When $p_{ui} \approx 0$, it only keeps initial node embeddings.

\nosection{Calculating User Preference}
Selecting a proper filter for each edge requires a priori knowledge of user preference.
However, user preference varies over time, e.g., a user may gradually lose interest in an item, decreasing conversion probability.
Therefore, \gnnname~calculates user preferences each time to support filter selection.

\textit{For unlabeled edges}, ConvE~\cite{ConvE}, an effective knowledge representation learning model, fits our scenario.
Since aggregated user and item embeddings in graph neural networks represent their general preferences and properties~\cite{NGCF,multi_behavior_GNN}, it can transform user embedding $e^{(l)}_{u}$ into preference $p_{ui}$ through item embedding $e^{(l)}_{i}$.
We define the preference calculation process as,
\begin{equation}
\begin{aligned}
    p_{ui} &= \mathrm{tanh}(c_{ui}),\\
    c_{ui} &= \mathrm{ConvE}(e^{(l)}_{u}, e^{(l)}_{i}) = \mathrm{Flatten}(\mathrm{CNN}(f(e^{(l)}_{u},e^{(l)}_{i})))W_{p},\nonumber
\end{aligned}
\end{equation}
where $f(\cdot)$ is a 2D reshaping and concatenation function, which converts $e^{(l)}_{u}$ and $e^{(l)}_{i}$ into two matrices and concatenates them row by row.
$\mathrm{CNN}$ denotes a 2D convolutional layer,
$\mathrm{Flatten(\cdot)}$ reshapes its output into a vector,
$W_{p}$ denotes a projection matrix, and $\mathrm{tanh}$ is the hyperbolic tangent function, which maps the result into $[-1, 1]$.

\textit{For positive edges}, we set $p_{ui} = 1$ as $e_{uiy}(t)$ is formed by the conversion between user $u$ and item $i$, i.e., the preference is maximized.
\section{EXPERIMENTS}
In this section, we conduct empirical studies on three large-scale datasets to answer the following four research questions. 
\textbf{RQ1}: How does \modelname~perform, compared with state-of-the-art methods for delayed feedback modeling (Section~\ref{subsec:RQ1})?
\textbf{RQ2}: Are the data pipeline and \gnnname~beneficial in \modelname~(Section~\ref{subsec:RQ2})?
\textbf{RQ3}: What are the impacts of different parameters on \modelname~(Section~\ref{subsec:RQ3})?
\textbf{RQ4}: How does \gnnname~deal with conversion and non-conversion relationships in our dynamic graph (Section~\ref{subsec:RQ4})?

\subsection{Experimental settings}
\nosection{Datasets}
We conduct experiments on three industrial datasets, i.e., \textit{TENCENT}, \textit{CRITEO2}, and \textit{CIKM}.
The statistics of these datasets are shown in Table~\ref{Table_Dataset_Statics}.
TENCENT is an industry dataset for social advertising competition\footnote{https://algo.qq.com/?lang=en}.
It includes 14 days of click and conversion logs.
We exclude data from the last two days.
CRITEO2 is a \textit{newly-released} industry dataset for the sponsored search of Criteo~\cite{CRITEO2018_DATASET}.
It represents the Criteo live traffic data in 90 days and publishes feature names additionally, \textit{which enables graph construction}.
CIKM is a competition dataset used in CIKM 2019 Ecomm AI\footnote{https://tianchi.aliyun.com/competition/entrance/231721/information?lang=en}.
It provides a sample of user behavior data in Taobao within 14 days.
We keep the first click and conversion for each pair of the user and item to adapt to the delayed feedback problem.

We divide all datasets into two parts evenly according to the click time of each sample.
We use the first part to pretrain CVR prediction models offline and divide the second part into multiple pieces by hours.
We train models on the $t$-th hour data and test them on the $(t+1)$-th data.
We reconstruct training data in the second part according to different data pipelines and remain evaluation data unchanged.
We report the \textit{average} performance across hours on the evaluation data.
Note that, among all datasets, \textit{CRITEO2} has the highest proportion of delayed conversions.

\begin{table*}[t]
\caption{Statistics of TENCENT, CRITEO, and CIKM dataset.}
\begin{tabular}{|c|c|c|c|c|c|c|c|}
\hline
\textbf{Datasets} & \textbf{Users}    & \textbf{Items}   & \textbf{Features} & \textbf{Samples}  & \textbf{Average CVR} & \textbf{Conversion in 0.25 Hours} & \textbf{Conversion in 24 Hours} \\ \hline
TENCENT   & 5,462,182  & 34,826   & 12       & 6,952,124  & 12.36\%      & 61.77\%                      & 92.83\%                       \\ \hline
CRITEO   & 10,437,254 & 1,628,064 & 10       & 12,167,296 & 10.51\%      & 29.54\%                      & 60.50\%                       \\ \hline
CIKM     & 977,886   & 1,953,421 & 8        & 9,710,244  & 10.84\%      & 42.45\%                      & 83.79\%                       \\ \hline
\end{tabular}
\label{Table_Dataset_Statics}
\end{table*}

\begin{table*}[t]
\caption{
Performance comparisons of \modelname~with state-of-the-art methods. 
}
\centering
\begin{tabular}{|c|l|l|l|l|l|l|}
\hline
\multirow{2}{*}{\textbf{Method}} & \multicolumn{2}{|c|}{\textbf{TENCENT}}                         & \multicolumn{2}{|c|}{\textbf{CRITEO2}}                         & \multicolumn{2}{|c|}{\textbf{CIKM}}                           \\ \cline{2-7} 
                        & \multicolumn{1}{|c|}{\textbf{\textit{AUC}}} & \multicolumn{1}{|c|}{\textbf{\textit{NILL}}} & \multicolumn{1}{|c|}{\textbf{\textit{AUC}}} & \multicolumn{1}{|c|}{\textbf{\textit{NILL}}} & \multicolumn{1}{|c|}{\textbf{\textit{AUC}}} & \multicolumn{1}{|c|}{\textbf{\textit{NILL}}} \\ \hline
PRETRAIN                & 75.07 $\pm$~0.38            & 32.57 $\pm$~0.20              & 70.92 $\pm$~0.45            & 29.81 $\pm$~0.14              & 67.70 $\pm$~0.13            & 30.71 $\pm$~0.05             \\
NODELAY                 & 78.00 $\pm$~0.04            & 31.05 $\pm$~0.02              & 71.97 $\pm$~0.04            & 29.23 $\pm$~0.01              & 68.54 $\pm$~0.18            & 30.12 $\pm$~0.04             \\ \hline
FSIW                    & 76.31 $\pm$~0.16            & 32.00 $\pm$~0.67              & 70.91 $\pm$~0.15            & 29.77 $\pm$~0.25              & 68.24 $\pm$~0.16            & 30.42 $\pm$~0.33             \\
FNW                     & 74.50 $\pm$~0.19            & 33.28 $\pm$~0.26              & 71.22 $\pm$~0.04            & 29.57 $\pm$~0.02              & 68.33 $\pm$~0.20            & 30.40 $\pm$~0.05             \\
ES-DFM                  & 76.32 $\pm$~0.20            & 31.98 $\pm$~0.12              & 71.45 $\pm$~0.05            & 29.68 $\pm$~0.03              & 68.38 $\pm$~0.17            & 30.39 $\pm$~0.21             \\
DEFER                   & 76.08 $\pm$~0.18            & 32.37 $\pm$~0.14              & 71.66 $\pm$~0.02            & 29.59 $\pm$~0.08              & 68.47 $\pm$~0.15            & 30.19 $\pm$~0.05             \\
DEFUSE                   & 76.25 $\pm$~0.17            & 32.15 $\pm$~0.16              & 71.85 $\pm$~0.06            & 29.52 $\pm$~0.10              & 69.58 $\pm$~0.19            & 30.23 $\pm$~0.13             \\ \hline
NGCF                    & 76.64 $\pm$~0.52            & 31.84 $\pm$~0.50              & \textbf{72.10 $\pm$~0.34}   & \textbf{29.40 $\pm$~0.12}     & 69.15 $\pm$~0.30            & 30.08 $\pm$~0.09             \\
FAGCN                   & 76.48 $\pm$~0.74            & 31.92 $\pm$~0.64              & 71.86 $\pm$~0.31            & 29.65 $\pm$~0.09              & 68.79 $\pm$~0.37            & 30.34 $\pm$~0.10             \\
TGN-attn                & \textbf{77.16 $\pm$~0.18}   & \textbf{31.75 $\pm$~0.10}     & 72.04 $\pm$~0.05            & 29.52 $\pm$~0.01              & \textbf{69.92 $\pm$~0.12}   & \textbf{29.95 $\pm$~0.03}    \\
\textbf{\modelname}                   & \textbf{77.82 $\pm$~0.17}   & \textbf{31.35 $\pm$~0.49}     & \textbf{73.44 $\pm$~0.06}   & \textbf{29.02 $\pm$~0.03}     & \textbf{71.49 $\pm$~0.15}   & \textbf{29.18 $\pm$~0.04}    \\ \hline
\%Improv.               & 93.86\%                & 80.47\%                  & 240.97\%                & 136.70\%                  & 449.70\%                & 260.64\%                 \\ \hline
\multicolumn{7}{l}{$^{\mathrm{*}}$In one column, the bold values correspond to the methods with the best and runner-up performances.}
\end{tabular}
\label{table:performance_comparison}
\end{table*}

\nosection{Comparison Methods}
We implement the following competitive methods as competitors for our proposed approach and categorize these methods into three classes.

\noindent \textbf{(1) Baselines}.
We implement two baseline methods, namely \textbf{PRETRAIN} and \textbf{NODELAY}.
PRETRAIN and NODELAY share the same model structure with DFM methods.
We train PRETRAIN offline and use it for online evaluation directly.
We train NODELAY with ground-truth labels and take its performance as the upper bound of all DFM methods as there is no delayed conversion.

\noindent \textbf{(2) Delayed Feedback Modeling (DFM)}.
We implement four delayed feedback modeling methods, including \textbf{FNW}~\cite{TWITTER_FNW}, \textbf{FSIW}~\cite{FSIW}, \textbf{ES-DFM}~\cite{ESDFM}, \textbf{DEFER}~\cite{DEFER}, and \textbf{DEFUSE}~\cite{DEFUSE}.
These methods are widely deployed in online commercial systems.
We use precisely the same data pipelines and loss functions as their original papers.

\noindent \textbf{(3) Graph Neural Network (GNN)}.
We implement three graph neural networks, including \textbf{NGCF}~\cite{NGCF}, \textbf{FAGCN}~\cite{high_low_filter_gnn}, and \textbf{TGN-attn}~\cite{TGN}. 
NGCF refines user and item representations via information from multi-hop neighbors. 
FAGCN uses high-pass and low-pass filters to model assortative and disassortative networks.
Since NGCF and FAGCN are \textit{static} graph neural networks, we build their graphs with offline training data and train their models with the online data stream.
TGN-attn, a \textit{dynamic} graph neural network, has the best performance among all variants of TGN.
We use positive samples to build the dynamic graph for TGN-attn, avoiding the damage of fake negatives.

\nosection{Evaluation Metrics}
We adopt two widely-used metrics for CVR evaluation, which describe model recommendation performance from different perspectives.

\noindent \textbf{(1) Area Under ROC Curve (AUC)}.
AUC~\cite{EXP_AUC_DEFINATION} is widely used in CVR predictions.
It measures the ranking performance between conversions and non-conversions; the higher, the better.

\noindent \textbf{(2) Negative Log Likelihood (NLL)}.
NLL~\cite{EXP_NLL_DEFINATION} is sensitive to the absolute value of CVR prediction.
It emphasizes the quality of the probabilistic model, and the lower, the better.

Following ESDFM~\cite{ESDFM} and DEFER~\cite{DEFER}, we compute relative improvement of our model over PRETRAIN as follows.
We take these results to demonstrate the performance of \modelname~straightforwardly; the higher, the better.
\begin{equation}
\begin{aligned}
\mathrm{\%Improv.=\frac{Metric_{\modelname}-Metric_{PRETRAIN}}{Metric_{NODELAY}-Metric_{PRETRAIN}}}.\nonumber
\end{aligned}
\end{equation}

\nosection{Experimental Settings}
We introduce experimental settings as follows.
We use Adam~\cite{Adam} as an optimizer with a learning rate of $0.0001$.
We set the batch size to $1,024$ for all competitors.
We repeat experiments ten times with different random seeds and report their average results.
Following ES-DFM~\cite{ESDFM}, we search time window length $l_{w}$ from $0.25$ hour.
We set the maximum time window length for searching to $24$ as it is the common attribution window length in many business scenarios~\cite{DEFER}.
We optimize all hyperparameters carefully through grid search for all methods to give a fair comparison.
The consumed resources vary with methods and hyperparameters. 

\subsection{Overall Performance Comparisons (RQ1)}
\label{subsec:RQ1} 
Table~\ref{table:performance_comparison} shows the comparison results on all datasets. 
From it, we observe that:
\textbf{(1)} In DFM methods, FNW and FSIW are inferior to ES-DFM as they either focus on data freshness or label accuracy.
ES-DFM, DEFER, and DEFUSE achieve superior performance in DFM methods on all datasets as they balance accuracy and freshness.
These results demonstrate that data freshness and label accuracy are essential for delayed feedback modeling.
\textbf{(2)} GNN methods generally outperform DFM methods as GNN can better represent user preferences and item properties.
Among GNN methods, NGCF surprisingly outperforms TGN-attn on \textit{CRITEO2}.
This result demonstrates that leveraging a dynamic graph does not always benefit delayed feedback modeling, as the samples used to build the dynamic graph are delayed and mislabeled.
\textbf{(3)} \modelname~is superior to both DFM and GNN methods.
It even outperforms NODELAY on \textit{CRITEO2} and \textit{CIKM}. 
These results demonstrate that our framework contributes to delayed feedback modeling as it obtains accurate conversion feedback and keeps the CVR prediction model fresh.

\subsection{Ablation Study (RQ2)}
\label{subsec:RQ2}
We verify the effectiveness of different components in \modelname~via ablation study.
We implement variants of our proposed method.
\textbf{(1)} \textit{Pretrain} and \textit{Oracle} are the lower and upper bounds of \modelname.
\textit{Pretrain} replaces the dynamic graph with a static graph built with offline training data.
\textit{Oracle} builds the dynamic graph and trains models with ground-truth labels.
\textbf{(2)} When building the dynamic graph, \textit{w/o Unlabel} removes unlabeled samples, while \textit{w/o POS} removes positive samples.
\textbf{(3)} \textit{GCN}~\cite{GCN} and \textit{GAT}~\cite{GAT} are two substitutes for~\gnnname.
\textit{w/o ConvE} replaces ConvE by two linear layers with Leaky ReLu~\cite{LeakyReLU} in \gnnname.
\textbf{(4)} \textit{w/o DD} replaces distribution debias with binary cross entropy loss.
From Table~\ref{TABLE_Ablation_Study}, we conclude:
\textbf{(1)} \modelname~largely outperforms \textit{Pretrain} and performs closer to \textit{Oracle}.
\textit{Pretrain} cannot capture frequent data distribution changes.
\textit{Oracle} degrades our data pipeline, \gnnname, and distribution debias, as they are specially designed for delayed feedback modeling, limiting the upper bound of our performance.
\textbf{(2)} For data pipelines, depending on positive or unlabeled samples alone cannot achieve the best recommendation performance.
The delivered positive samples are critical in the E-commerce scenario, i.e., \textit{CIKM}, as positive behaviors strongly reflect user preference.
In contrast, the delivered unlabeled samples are necessary for the scenario with highly delayed conversions, i.e., \textit{CRITEO2}.
\textbf{(3)} For \gnnname, high-pass and low-pass filters are indispensable to achieve ideal performance.
The underperformance of \textit{GAT} and \textit{GCN} demonstrates that conventional graph neural networks are unsuitable for our dynamic graph, as they only use low-pass filters.
The difference between \textit{w/o ConvE} and \modelname~demonstrates that $\mathrm{ConvE}$ can better calculate the user preference for the item.
\textbf{(4)} The result of \textit{w/o DD} shows that our distribution debias benefits model performance, as it corrects the distribution shifts.
\textbf{(5)} All components in \modelname~are essential and our proposed data pipeline is the most crucial part for delayed feedback modeling.

\begin{table}[b]
\caption{Performance of all variants of \modelname.}
\begin{tabular}{|c|l|l|l|l|l|l|}
\hline
\multirow{2}{*}{\textbf{Method}} & \multicolumn{2}{|c|}{\textbf{TENCENT}}                         & \multicolumn{2}{|c|}{\textbf{CRITEO2}}                         & \multicolumn{2}{|c|}{\textbf{CIKM}}                           \\ \cline{2-7} 
                        & \multicolumn{1}{|c|}{\textbf{\textit{AUC}}} & \multicolumn{1}{|c|}{\textbf{\textit{NILL}}} & \multicolumn{1}{|c|}{\textbf{\textit{AUC}}} & \multicolumn{1}{|c|}{\textbf{\textit{NILL}}} & \multicolumn{1}{|c|}{\textbf{\textit{AUC}}} & \multicolumn{1}{|c|}{\textbf{\textit{NILL}}} \\ \hline
Pretrain                   & 76.70                   & 31.87                    & 72.16                   & 29.30                    & 69.44                   & 30.10                    \\
\multicolumn{1}{|c|}{Oracle} & 78.72                   & 30.70                    & 73.78                   & 28.88                    & 72.75                   & 28.73                    \\ \hline
w/o Unlabel                        & 77.50                   & 31.79                    & 72.74                   & 29.18                    & 70.23                   & 29.70                    \\
w/o POS                  & 77.29                   & 31.67                    & 73.22                   & 29.11                    & 67.93                   & 30.22                    \\ \hline
GCN                        & 76.98                   & 31.58                    & 73.28                   & 29.05                    & 71.01                   & 29.51                    \\
GAT                        & 77.53                   & 31.71                    & 72.37                   & 31.18                    & 70.50                   & 29.73                    \\ 
w/o ConvE                        & 77.67                   & 31.37                    & 73.31                   & 29.13                    & 71.32                   & 29.33                    \\ \hline
w/o DD                        & 77.38                   & 31.56                    & 72.26                   & 29.72                    & 69.53                   & 30.58                    \\ \hline
\textbf{\modelname}             & \textbf{77.82}          & \textbf{31.35}           & \textbf{73.44}          & \textbf{29.02}           & \textbf{71.49}          & \textbf{29.18}           \\ \hline
\end{tabular}
\label{TABLE_Ablation_Study}
\end{table}

\begin{figure*}[t]
\subfigure[Time window length $l_{w}$.]{
\includegraphics[width=4.2cm]{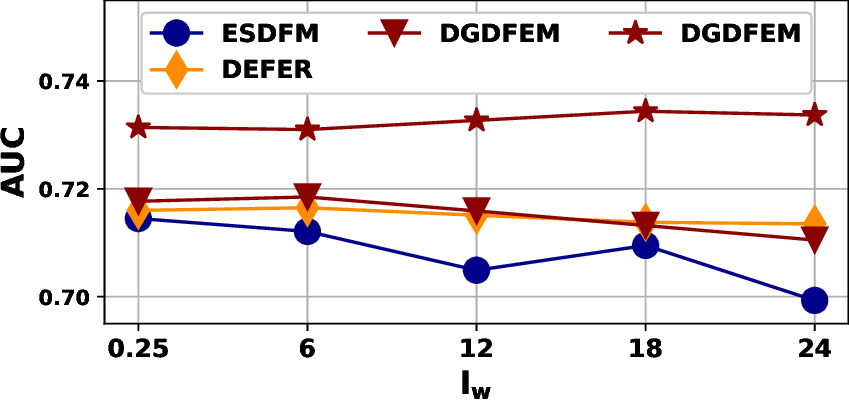}
\label{fig:performance with different lw}
}
\subfigure[Hop numbers $k$.]{
\includegraphics[width=4.2cm]{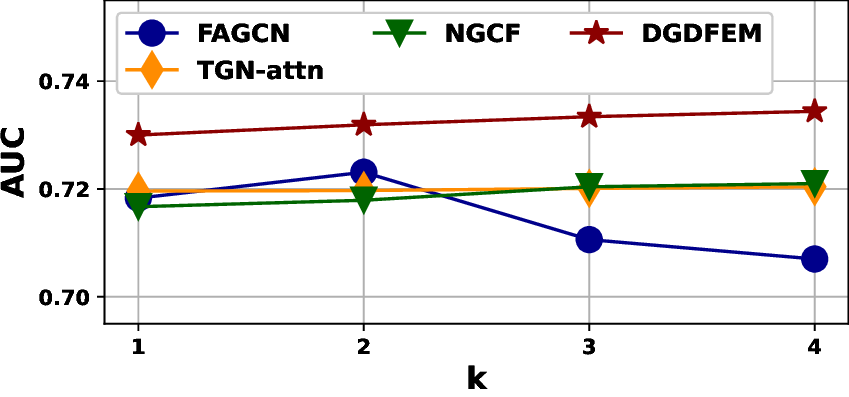}
\label{fig:performance with different K}
}
\subfigure[Neighbor Numbers $m$.]{
\includegraphics[width=4.2cm]{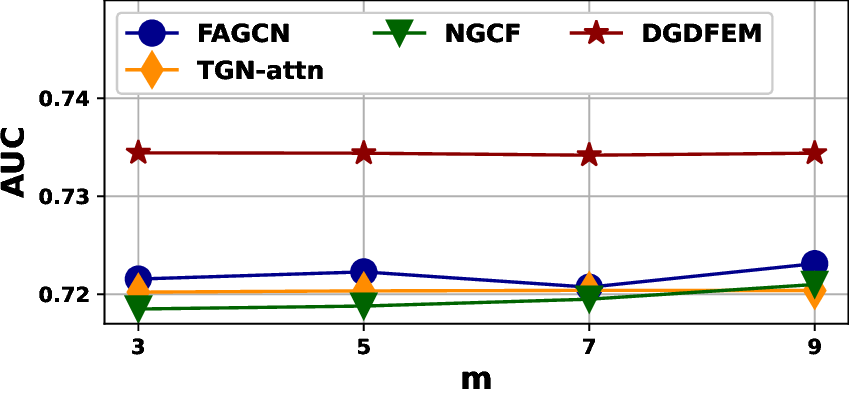}
\label{fig:performance with different M}
}
\subfigure[Embedding retention proportion $\epsilon$.]{
\includegraphics[width=4.2cm]{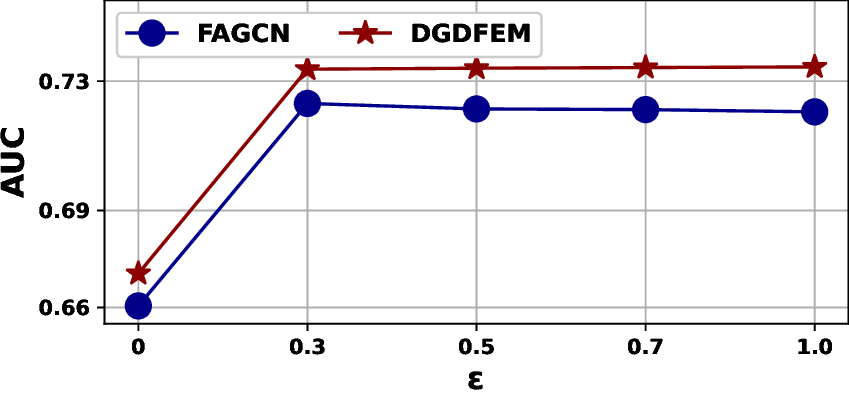}
\label{fig:performance with different e}
}
\caption{Performance with different Parameters.}
\label{fig:performance with different parameters}
\end{figure*}

\begin{figure*}[t]
\subfigure[Variations on \textit{TENCENT}]{
\includegraphics[width=5.5cm]{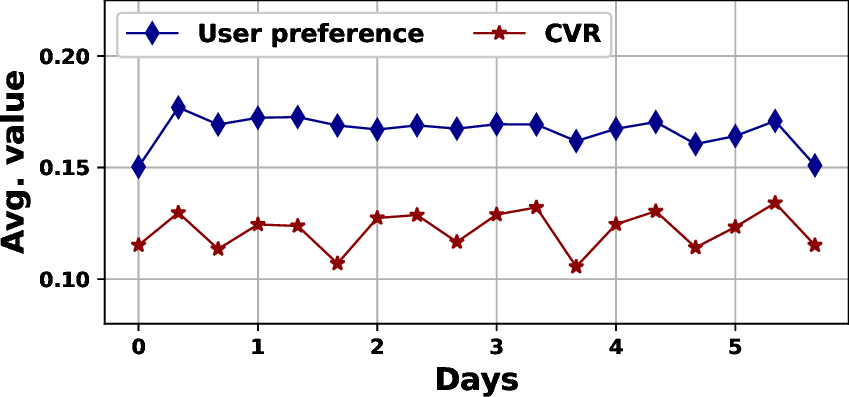}
\label{figure_case_study_tecent}
}
\subfigure[Variations on \textit{CRITEO2}]{
\includegraphics[width=5.5cm]{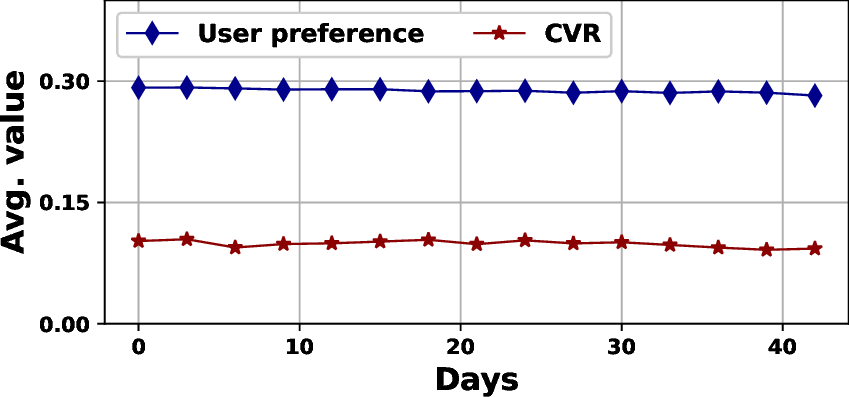}
\label{figure_case_study_criteo}
}
\subfigure[Variations on \textit{CIKM}]{
\includegraphics[width=5.5cm]{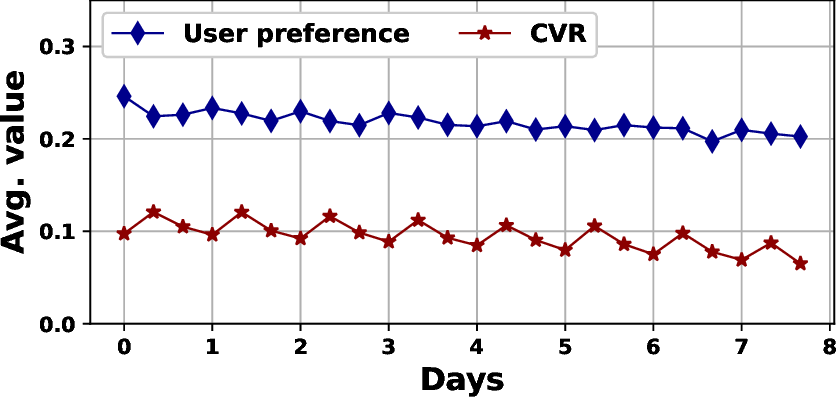}
\label{figure_case_study_cikm}
}
\caption{Filter and CVR variations with time elapse}
\label{figure_case_study}
\end{figure*}

\subsection{Parameter Analyses (RQ3)}
\label{subsec:RQ3}
We evaluate the impacts of different model parameters on \modelname~and state-of-the-art methods with \textit{CRITEO2}.
\textit{CRITEO2} best reflects the delayed feedback problem as it has the highest proportion of delayed conversions.
From Fig.~\ref{fig:performance with different parameters}, we conclude:
\textbf{(1)} For time window length $l_{w}$, all methods except \modelname~has performance degradation as $l_{w}$ increases.
This result demonstrates that \modelname~successfully solves the trade-off between data freshness and label accuracy.
In \modelname, we can either set high $l_{w}$ to increase model performance or use low $l_{w}$ to save resources. 
\textbf{(2)} For hop number $k$, the performance of all methods except FAGCN slightly improves as $k$ increases.
This indicates that signals from deeper neighbors are less beneficial for delayed feedback modeling than other scenarios.
\textbf{(3)} For neighbor number $m$, the performance of \modelname, TGN-attn, and FAGCN has no significant relationship with $m$, whereas the performance of NGCF slightly improves as $m$ increases.
This indicates that it is unnecessary to set large $m$ for delayed feedback scenarios. 
\textbf{(4)} For retention proportion $\epsilon$, \modelname~and FAGCN perform much worse with $\epsilon = 0$, indicating that only depending on neighbor features is insufficient.
\modelname~prefers a higher retention proportion of initial node embeddings, whereas FAGCN prefers a lower one.
Our proposed \gnnname~is intrinsically different from FAGCN, although they both use high-pass and low-pass filters.
FAGCN aims to capture node differences or retain node commonalities among conversion relationships.
In contrast, \gnnname~aims to deal with conversion and non-conversion relationships in the dynamic graph.

\subsection{Case Study (RQ4)}
\label{subsec:RQ4}
As motivated in Section~\ref{subsec:AHLPF}, \gnnname~should use a low-pass filter for an edge with high conversion possibility, and vice versa.
It calculates user preferences to \textit{directly} determine the filter weight and type (see~\eqref{Equation_AHLPF}).
Thus, the CVR of the system should correlate with the average user preferences when time elapses.
To illustate it clearly, we use a sliding window with different lengths to average CVR and filter weight in the window.
We set the window length as 8 hours for both \textit{TENCENT} and \textit{CIKM}, and 72 hours for \textit{CRITEO2}.
From Fig.~\ref{figure_case_study}, we can observe that the average user preferences correlate with the CVR.
This phenomenon is particularly significant on \textit{TENCENT} and \textit{CIKM}.
%
These results demonstrate that \gnnname~can use correct filters to deal with non-conversion and conversion relationships.
\section{Conclusion}
As the first work to exploit graph neural network in delayed feedback modeling, \modelname~successfully solves the trade-off between label accuracy and data freshness.
The framework of \modelname~includes three stages, i.e., preparing the data pipeline, building the dynamic graph, and training the CVR prediction model.
We further propose a novel graph convolutional method named \gnnname.
The dynamic graph captures frequent data distribution changes and \gnnname~leverages appropriate filters for the edges in the graph.
We conduct extensive experiments on three large-scale industry datasets and the results validate the success of our framework.

\section*{Acknowledgment}
This work was supported in part by the National Key R\&D Program of China (No. 2022YFF0902704) and the ”Ten Thousand Talents Program” of Zhejiang Province for Leading Experts (No. 2021R52001).

\bibliographystyle{IEEEtran}
\bibliography{reference}
\newpage

\appendix
\section{Proof}
\subsection{Proof of Proposition 1}
\label{appendix_proof_proposition1}
\begin{IEEEproof}
In this section, we prove that distribution debias can correct distribution shifts.
We let $p(y|x)$ and $b(y|x)$ denote the actual data distribution and biased distribution respectively, where $x$ denotes sample features and $y$ denotes sample labels. 
Our proof includes two parts, i.e., bridging two distributions and leveraging importance sampling.

\nosection{Bridging Two Distributions} Let us first suppose a scenario, where we do not calibrate fake negatives as positive for model training.
In this scenario, the relationship between two distributions is,
\begin{equation}
\begin{aligned}
b(y=0|x) &= 1-p(y=1|x)+p(y=1|x)p(t_{d}>l_{w}|x,y=1),\\
b(y=1|x) &= p(y=1|x)p(t_{d}\leq l_{w}|x,y=1).\nonumber
\end{aligned}
\end{equation}

In our data pipeline, we calibrate fake negatives as positive and \textit{deliver these samples again.}
The total number of samples are increased by $p(y=1|x)p(t_{d}>l_{w}|x,y=1)$.
We have
\begin{small}
\begin{equation}
\begin{aligned}
b(y=0|x) &= \frac{1-p(y=1|x)+p(y=1|x)p(t_{d}>l_{w}|x,y=1)}{1+p(y=1|x)p(t_{d}>l_{w}|x,y=1)},\\
b(y=1|x) &= \frac{p(y=1|x)(p(t_{d}\leq l_{w}|x,y=1)+p(t_{d}>l_{w}|x,y=1))}{1+p(y=1|x)p(t_{d}>l_{w}|x,y=1)}\\
&= \frac{p(y=1|x)}{1+p(y=1|x)p(t_{d}>l_{w}|x,y=1)}.
\end{aligned}
\label{equ_two_distribution_relationship}
\end{equation}
\end{small}

Thus, we build the relationship between actual data distribution and biased distribution.
\nosection{Leveraging Importance Sampling}
We define the ideal loss for model optimization in true data distribution $p(y|x)$ and actual loss in biased data distribution $b(y|x)$ as $\mathcal{L}_{ideal}$ and $\mathcal{L}_{true}$, respectively.
We let $f_{\theta}$ denote the CVR prediction model. %
Following the importance sampling~\cite{importance_sampling}, we obtain an unbiased estimate of the true data distribution as,
\begin{equation}
\begin{aligned}
\mathcal{L}_{ideal}(\theta)&=-\mathbb{E}_{p}[logf_{\theta}(y|x)]\\
&=-\sum_{x,y}p(y|x)logf_{\theta}(y|x)\\
&=-\sum_{x,y}\frac{p(y|x)}{b(y|x)}b(y|x)logf_{\theta}(y|x)\\
&\approx -\mathbb{E}_{b}[\frac{p(y|x)}{b(y|x)} logf_{\theta}(y|x)].
\end{aligned}
\end{equation}

If we have fully trained the prediction models, we can directly take the values of $\hat{y}_{p}$ and $\hat{y}_{fn}$ as $p(y=1|x)p(t_{d}\leq l_{w}|x,y=1)$ and $p(y=1|x)p(t_{d}>l_{w}|x,y=1)$ in~\eqref{equ_calculating_three_equations}.
Combining with~\eqref{equ_two_distribution_relationship}, we calculate $\frac{p(y|x)}{b(y|x)}$ as,
\begin{equation}
\begin{aligned}
\frac{p(y=0|x)}{b(y=0|x)} &= (1+\hat{y}_{fn})(1-\hat{y}_{p}-\hat{y}_{fn}),\\
\frac{p(y=1|x)}{b(y=1|x)} &= 1+\hat{y}_{fn}.\nonumber
\end{aligned}
\end{equation}

Then, we can replace these parts in~\eqref{equ_loss_function} accordingly. 
We have
\begin{small} 
\begin{equation}
\begin{aligned}
\mathcal{L} &= \sum y_{ui}\frac{p(y=1|x)}{b(y=1|x)}log(\hat{y}_{cvr})+(1-y_{ui})\frac{p(y=0|x)}{b(y=0|x)}log(1-\hat{y}_{cvr})\\
&=-\mathbb{E}_{b}[\frac{p(y|x)}{b(y|x)} logf_{\theta}(y|x)]\\
&=\mathcal{L}_{ideal}(\theta),\nonumber
\end{aligned}
\end{equation}
\end{small} 
which concludes that our distribution debias corrects distribution shifts.
\end{IEEEproof}

\subsection{Time Complexity}
\label{Time Complexity}
Here, we mainly discuss the time complexity of \gnnname~ since it is the most time-consuming operation in \modelname.
It involves calculating preferences and aggregating features.
Let $n_{e}$, $\hat{n}_{e}$, and $t$ denote the edge number, unlabeled edge number, and time complexity for ConvE, respectively. 
The former has computational complexity $\mathcal{O}(t\hat{n}_{e})$.
The latter has the computational complexity $\mathcal{O}(n_{e})$.
Thus, the overall time complexity for $k$ layers of \gnnname~is $\mathcal{O}(kn_{e}+kt\hat{n}_{e})$.
Since \modelname~limits neighbor number $m$ and hop number $k$, the upper bounds of $n_{e}$ and $\hat{n}_{e}$ are determined as $(1-m^k)/(1-m)$.
Thus, the time complexity for \gnnname~is bounded, proving that we can apply \modelname~into online commercial systems.

\section{Implementation details}
During optimization, the weight of each sample requires accurate predictions of $\hat{y}_{fn}$ and $\hat{y}_{p}$.
These two prediction models are not fully trained at first, so their prediction values are initially incorrect.
Some methods leverage normalization on the prediction values to stabilize model training~\cite{normalize_motivation, normalize_recommendation1,normalize_recommendation2}.
However, in our scenario, $\hat{y}_{fn}$ and $\hat{y}_{p}$ should represent the actual percentages of \textit{fake negatives} and \textit{positive} in the true data distribution. 
Normalization alters the mean and standard deviation of $\hat{y}_{fn}$ and $\hat{y}_{p}$, which leads to a wrong relationship between the true and biased data distribution.
Here, we first train the CVR prediction model offline with average binary cross-entropy loss to give accurate predictions of $\hat{y}_{fn}$ and $\hat{y}_{p}$.
We then replace it with our proposed loss function in~\eqref{equ_loss_function} during online training to correct distribution shifts.

\end{document}